\documentclass[runningheads]{llncs}
\usepackage{amsmath}
\usepackage{amsfonts}
\usepackage{amssymb}
\usepackage{graphicx}
\usepackage{url,hyperref}
\usepackage{algorithmic}
\usepackage{algorithm}%
\usepackage[T1]{fontenc}

\begin{document}

\title{Attention-Based Estimation of the Individual Treatment Benefit Probability under
Dose Variation}
%
%
\author{Lev V. Utkin\inst{} 
\and Andrei V. Konstantinov\inst{} 
\and Stanislav K. Kogan\inst{} 
\and Natalya M. Verbova\inst{}
\and Maksim I. Goriunov\inst{}}

\authorrunning{L.V. Utkin et al.}
%
\institute{Peter the Great St.Petersburg Polytechnic University \\
Higher School of Artificial Intelligence Technologies \\
St.Petersburg, Russia\\
\email{utkin\_lv@spbstu.ru, konstantinov\_av@spbstu.ru, kogan\_sk@spbstu.ru, verbova\_nm@spbstu.ru, goryunov\_myu@spbstu.ru}}%
\maketitle       

\begin{abstract}
Estimating the probability that a treatment outperforms a control for an
individual patient, called the Individual Probability of Treatment Benefit
(IPTB), offers a clinically intuitive alternative to population-average
metrics. However, existing methods for IPTB estimation are largely confined to
binary treatment settings, despite the prevalence of dose-varying
interventions in clinical practice. We propose a general framework for IPTB
estimation with ordinal outcomes under discrete dose assignments, called
Dose-AIPTB (Dose Attention-based IPTB). Our approach recasts the problem as
binary classification over the unobserved sign of the individual treatment
effect, constructing pseudo-labels from covariate-similar pairwise comparisons
and aggregating them via attention mechanisms or Nadaraya-Watson kernel
regression. This formulation naturally accommodates multiple discrete dose
levels, extending beyond the binary treatment paradigm. Through numerical
experiments on real-world and synthetic data under covariate shift, varying
sample sizes, and heterogeneous outcomes, we demonstrate that attention-based
aggregation consistently outperforms kernel alternatives. The framework
provides a foundation for personalized dose selection grounded in
individual-level benefit probabilities. Codes implementing the model are
publicly available at \url{https://github.com/NTAILab/AIPTBDose}.


\end{abstract}

\section{Introduction}

Clinical decision-making has long been guided by evidence generated at the
population level. The \emph{Average Treatment Effect} (ATE) stands as the
cornerstone of this paradigm, offering a singular measure that determines
whether a therapy outperforms a control for a prototypical patient. This
metric has proven indispensable for regulatory approval and the formulation of
clinical guidelines. Yet, a fundamental tension arises when the ATE is
transported from the research setting to the bedside. Whereas the ATE resolves
uncertainty about group-level superiority, the clinician faces a different
question: \emph{Will this specific patient, with this particular constellation
of characteristics, derive benefit?} The principle of clinical equipoise,
which justifies randomization at the trial level, offers little guidance for
the individual case.

This disconnect has spurred a shift toward individualization. The Conditional
Average Treatment Effect (CATE) emerged as a natural extension, aiming to
characterize how treatment effects vary across subgroups defined by patient
covariates. However, even CATE, which estimates the \emph{expected} difference
in outcomes, falls short of addressing the clinician's core concern. What
matters in practice is not merely the expected magnitude of benefit, but the
probability that a patient will experience a favorable outcome relative to the
alternative. This motivates the \emph{Individual Probability of Treatment
Benefit} (IPTB), defined for a patient with covariates $X=\mathbf{x}$ as
\begin{equation}
\rho(\mathbf{x})=\Pr \{H>Y\mid X=\mathbf{x}\}=\Pr \{ \Delta>0\mid X=\mathbf{x}%
\},
\end{equation}
where $H$ and $Y$ denote the potential outcomes under treatment and control,
respectively, and $\Delta=H-Y$ captures the individual-level treatment effect.
Unlike CATE, which summarizes the distribution of $\Delta$ through its first
moment, the IPTB directly targets the probability of a positive outcome as a
quantity that aligns directly with the notion of treatment benefit in clinical reasoning.

The appeal of the IPTB belies its statistical complexity. Estimating a mean
effect (whether unconditional (ATE) or conditional (CATE)) requires only the
ability to model the conditional expectation of the observed outcome. By
contrast, estimating $\Pr \{ \Delta>0\mid X=\mathbf{x}\}$ demands recovering
the full conditional distribution of potential outcomes, or at least the joint
distribution of $(H,Y)$ given covariates. This task is fundamentally
complicated by the central obstacle of causal inference: for any given
individual, at most one potential outcome is ever observed. The counterfactual
outcome remains hidden, making the sign of $\Delta$ inherently unobservable at
the individual level.

Reconstructing the distribution of $\Delta$ thus requires either parametric
assumptions about the underlying data-generating process or sophisticated
nonparametric strategies capable of imputing the missing counterfactual
distribution. Bayesian approaches naturally accommodate this challenge by
modeling joint potential outcome distributions \cite{Zhang-Zhang-etal-22a},
while frequentist methods have increasingly turned to techniques such as
probabilistic classification or direct modeling of the benefit function
\cite{Melnychuk-etal-22,melnychuk2024quantifying}. What unifies these efforts
is the recognition that inferring individual benefit probabilities imposes
stronger demands both in terms of assumptions and computational complexity
than conventional effect estimation.

The literature on heterogeneous treatment effect estimation has witnessed
substantial methodological evolution. Early approaches relied on regularized
regression with interaction terms \cite{Jeng-Lu-Peng-2018}. Subsequent
advances introduced meta-learners, including the T-learner, S-learner,
X-learner, and DR-learner \cite{kennedy2023towards}, that combine base
estimators in flexible ways to estimate CATE
\cite{Kunzel-etal-19,salditt2024tutorial,Wang-etal-2016}. Deep learning
architectures have since been deployed to capture complex, high-dimensional
relationships in covariate spaces \cite{Curth-Schaar-21a,Nair-etal-22,Qin-Wang-Zhou-21,shi2024estimating}. Nonparametric kernel methods
\cite{Imbens-04,Park-Shalit-etal-21} and transformer-based attention
mechanisms \cite{Guo-Zheng-etal-21,Melnychuk-etal-22,Zhang-Zhang-etal-22}
represent more recent frontiers, offering flexible function approximation with
varying trade-offs in interpretability and scalability.

Parallel developments have targeted IPTB estimation specifically. These
include Bayesian frameworks that model the joint distribution of potential
outcomes \cite{Zhang-Zhang-etal-22a}, as well as approaches that reframe the
problem as probabilistic classification, where the target is the probability
that the treated outcome exceeds the control outcome \cite{Melnychuk-etal-22,melnychuk2024quantifying}. Despite these advances, a common limitation
persists across much of this literature: the implicit assumption of binary
treatment assignment.

In many clinical contexts, treatment is not simply present or absent; it is
administered in varying intensities, frequencies, or quantities. Whether
conceptualized as the number of drug units, the duration of therapy, or the
concentration of an active ingredient, the \emph{dose} introduces a layer of
complexity that binary formulations cannot accommodate. A patient may derive
benefit from a low dose, experience toxicity at a high dose, or exhibit a
non-monotonic response that defies simple dose response assumptions. Moreover,
the optimal dose defined as the minimum level sufficient to achieve a positive
treatment effect may vary across individuals in ways that reflect underlying
biological heterogeneity.

This gap has not gone unnoticed. Recent work has begun exploring causal
inference with complex treatments, including continuous and multi-valued
interventions \cite{Hirzli-etal-24,parikh2024safe,piskorz2025beyond,schroder2025conformal,wang2026causal}. The problem of multiple treatment
versions has been examined from a causal identification perspective
\cite{beal2020causal,vanderweele2013causal}. Notably, neural network-based
approaches have been proposed for estimating individual dose response curves
across continuous dosage parameters \cite{Schwab-etal-2020}. However, these
methods typically target expected outcomes or dose response surfaces rather
than the probability of benefit a quantity that may better capture clinically
meaningful thresholds of efficacy.

In this work, we introduce a general framework for estimating the IPTB with
continuous or ordinal outcomes in settings where treatment is administered at
discrete dose levels. It is called Dose-AIPTB (Dose Attention-based IPTB). Our
approach recasts the estimation problem as binary classification, where the
target label is the sign of the individual treatment effect $\Delta$. Since
$\Delta$ is unobservable, we construct pseudo-labels from pairwise comparisons
between treated and control patients with similar covariate profiles. These
comparisons are aggregated using feature-dependent weighting mechanisms that
respect covariate similarity: we implement two complementary strategies based
on Nadaraya-Watson kernel regression \cite{Nadaraya-1964,Watson-1964} and
dot-product attention \cite{Luong-etal-2015,Vaswani-etal-17}. The resulting
framework inherits theoretical grounding from recent advances in
distributional causal inference \cite{Konstantinov-Utkin-etal-25}.

Our focus on discrete dose settings, exemplified by scenarios such as the
number of drugs co-administered, addresses a practically relevant yet
methodologically underexplored domain. For a given patient, a lower dose may
confer benefit while a higher dose proves detrimental, and the optimal dose
likely depends on patient-specific characteristics. By directly modeling the
probability that treatment benefit exceeds a clinically meaningful threshold
(here defined as superiority over control), our approach provides a foundation
for dose personalization grounded in probabilistic benefit estimates.

The contributions of this work are threefold:

\begin{enumerate}
\item We propose a nonparametric model for IPTB estimation that avoids
parametric distributional assumptions. The attention-based aggregation
mechanism enables flexible borrowing of information from similar historical
cases. The main idea behind the model is to consider all pairs of instances
such that one instance in the pair is from the control group and another
instance is from the treatment group. Considering pairs significantly
increases the training sample and allows us to validate the model for real data.

\item Our framework explicitly accommodates varying treatment doses, moving
beyond the binary treatment paradigm that dominates existing IPTB literature.

\item We evaluate the proposed approach through comprehensive numerical
experiments on both real-world and synthetic datasets, examining performance
under covariate shift, varying sample sizes, and heterogeneous outcome
structures. The proposed model is compared with the well-known meta-learners:
T-learner, S-learner, X-learner, and DR-learner. Random forests
\cite{Breiman-2001} are selected as the base learners for all meta-learners.
All codes are publicly available at \url{https://github.com/NTAILab/AIPTBDose}.
\end{enumerate}

\section{Problem Formulation}

This work considers an observational study comprising two distinct cohorts: a
control group and a treatment group. The control cohort is represented by the
dataset $\mathcal{D}_{0} = \{ (\mathbf{x}_{i}, y_{i}) \}_{i=1}^{c}$,
containing $c$ independent observations. For each subject $i$, $\mathbf{x}_{i}
\in \mathbb{R}^{d}$ denotes a $d$-dimensional covariate vector, and $y_{i}
\in \mathbb{R}$ represents the observed outcome under the control condition
(e.g., survival time or a physiological measure).

Similarly, the treatment cohort is denoted by $\mathcal{D}_{1}=\{(\mathbf{z}%
_{j},h_{j},A)\}_{j=1}^{t}$, consisting of $t$ subjects. Here, $\mathbf{z}%
_{j}\in \mathbb{R}^{d}$ is the covariate vector, and $h_{j}\in \mathbb{R}$ is
the observed outcome following the intervention. To formalize the study
design, we define a treatment dosage $A\in \{1,...,m\}$. Generally, the
parameter $A$ can also be added to the control patients assuming the condition
$A=0$. However, this is not necessary in the context of the proposed model.

Adopting the potential outcomes framework, let $Y$ and $H$ denote the
counterfactual outcomes for a subject under control and treatment ($A\geq1$),
respectively. Unlike conventional approaches that target the Conditional
Average Treatment Effect (CATE), our inference focuses on the probability of
individual benefit. Concretely, for a subject with covariate vector
$\mathbf{X}=\mathbf{x}$, we seek to estimate:
\begin{equation}
\Psi(\mathbf{x})=\Pr \left(  H>Y\mid \mathbf{X}=\mathbf{x}\right)  .
\end{equation}
This quantity offers a probabilistic assessment of treatment efficacy at the
individual level, moving beyond an expectation of the outcome difference to
capture the likelihood that treatment proves superior to control for a given patient.

\section{Treatment Benefit Probability as a Classification Task}

Three main ideas behind the proposed models are the following: (1) the
pairwise patient comparison; (2) treatment doses as an additional feature in
the vector of features for treatment patients; (3) the attention mechanism for
computing probability that the difference of outcomes for a pair of patients
from the control and treatment groups is positive.

\subsection{Pairwise patient comparison}

The core methodology involves constructing pairwise comparisons between
patients from distinct cohorts. Specifically, we form pairs consisting of one
treatment group patient $(\mathbf{z}_{i},h_{i})$ and one control group patient
$(\mathbf{x}_{j},y_{j})$. The treatment effect for each pair is quantified as
$\Delta_{ij}=h_{i}-y_{j}$.

We collect all ordered treatment effect values $\Delta_{ij}$ for $i=1,\dots,t$
and $j=1,\dots,c$, into two subsets $\mathcal{G}^{+}$ and $\mathcal{G}^{-}$
with positive and negative values $\Delta_{ij}$, respectively, i.e., we can
write
\begin{equation}
\mathcal{G}^{+}=\{ \Delta_{ij}:\Delta_{ij}>0\},\  \mathcal{G}^{-}=\{
\Delta_{ij}:\Delta_{ij}\leq0\}.
\end{equation}

Let us consider subsets of $\Delta_{ij}$ whose values can belong to the
interval $(0,+\infty)$, and introduce the index set $\mathcal{R}^{+}$
corresponding to all pairs of observations with positive values $\Delta_{ij}$
(from $\mathcal{G}^{+}$):
\begin{equation}
\mathcal{R}^{+}=\{(i,j):\Delta_{ij}>0\}.
\end{equation}

This case is simple because the probability that $\Delta_{ij}>0$ is $1$,
i.e.,
\begin{equation}
\Pr \{ \Delta>0\mid \mathbf{Z}=\mathbf{z}_{i},\mathbf{X}=\mathbf{x}%
_{j}\}=1,\ (i,j)\in \mathcal{R}^{+}.
\end{equation}

We also introduce the index set $\mathcal{R}^{-}$ such that
\begin{equation}
\mathcal{R}^{-}=\{(i,j):\Delta_{ij}\leq0\}.
\end{equation}

\begin{figure}
[ptb]
\begin{center}
\includegraphics[
height=1.0in,
width=5.0in
]%
{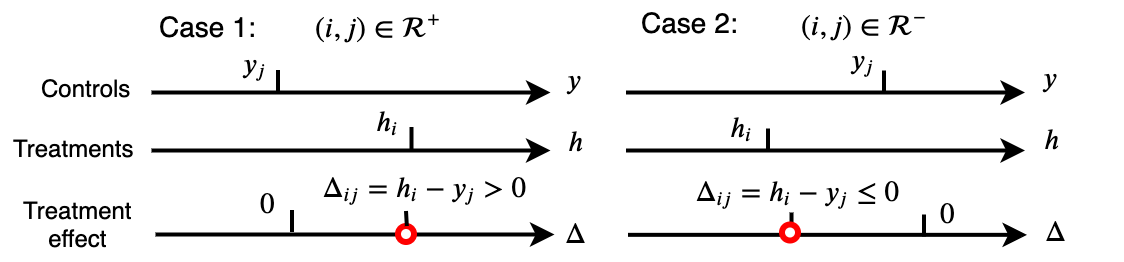}%
\caption{Two cases of differences $\Delta_{ij}$}%
\label{f:explan_iptb_simpl}%
\end{center}
\end{figure}

The corresponding cases of the differences $\Delta_{ij}=h_{i}-y_{j}$ are
depicted in Fig. \ref{f:explan_iptb_simpl}.%

Our aim is find the probability that the treatment effect will be positive for
a new pair of patients with feature vectors $\mathbf{z}$ and $\mathbf{x}$,
i.e.,
\[
p^{+}(\mathbf{z},\mathbf{x})=\Pr \{ \Delta>0\mid \mathbf{Z}=\mathbf{z}%
,\mathbf{X}=\mathbf{x}\}.
\]

In other words, $p^{+}(\mathbf{z},\mathbf{x})$ is a probability that the
difference of observations with the treatment instance $\mathbf{z}$ and the
control instance $\mathbf{x}$ produces $\Delta_{ij}$ which belongs to the set
$\mathcal{R}^{+}$.

\subsection{Treatment doses}

In order to take into account different treatment doses, it is proposed to add
this quantity to feature vectors of the treatment patients. At that, there are
different types of the treatment doses: discrete, continuous, and categorical.
Discrete and continuous types are represented by an additional feature whereas
the categorical type is represented by a set of additional features using the
one-hot coding scheme.

We consider the first discrete type of the treatment dosage to simplify the
description. It is important to note that the consideration of the treatment
dose as an additional feature in the concatenated vectors $\mathbf{z}$ and
$\mathbf{x}$ requires to weigh this feature and other features when the
dimension of the feature vectors is rather large. In this case, the weight
$\omega$ is assigned to the dose feature and weights $(1-\omega)/(2d)$ are
assigned to other features. As a result, distances between feature vectors are
calculated in accordance with the weights.

\subsection{Attention for solving the classification problem}

Let us consider the probability $p^{+}(\mathbf{z},\mathbf{x})$ from the
attention mechanism or from the kernel Nadaraya-Watson regression point of
view. We aim to find this probability of the positive treatment effect for the
pair of new patients with the feature vectors $\mathbf{z}$ and $\mathbf{x}$.
For every pair of vectors $\mathbf{z}$ and $\mathbf{x}$, we find the
concatenations $(\mathbf{z},\mathbf{x})$ and $(\mathbf{z}_{i},\mathbf{x}_{j}%
)$. Then the Nadaraya-Watson kernel regression can be written as follows:
\begin{align}
p^{+}(\mathbf{z},\mathbf{x}) &  =\sum_{(l,k)\in \mathcal{R}^{+}\cup
\mathcal{R}^{-}}a((\mathbf{z},\mathbf{x)},(\mathbf{z}_{l},\mathbf{x}%
_{k}))\cdot \mathbf{1}[\Delta_{lk}>0]\nonumber \\
&  =\sum_{(i,j)\in \mathcal{R}^{+}}a((\mathbf{z},\mathbf{x)},(\mathbf{z}%
_{i},\mathbf{x}_{j}))\cdot1+\sum_{(r,s)\in \mathcal{R}^{-}}a((\mathbf{z}%
,\mathbf{x)},(\mathbf{z}_{r},\mathbf{x}_{s}))\cdot0,
\end{align}
where the attention weight $a((\mathbf{z},\mathbf{x)},(\mathbf{z}%
_{i},\mathbf{x}_{j}))$ conforms with relevance of the concatenated feature
vector $(\mathbf{z},\mathbf{x)}$ to the concatenated feature vector
$(\mathbf{z}_{i},\mathbf{x}_{j})$, and there holds%
\begin{equation}
\sum_{(r,s)\in \mathcal{R}^{+}\cup \mathcal{R}^{-}}a((\mathbf{z},\mathbf{x)}%
,(\mathbf{z}_{r},\mathbf{x}_{s}))=1.
\end{equation}

It can be seen from the above that Nadaraya-Watson regression model estimates
$p^{+}(\mathbf{z},\mathbf{x})$ as a weighted sum of training outputs, which
are indicator functions $\mathbf{1}[\Delta_{ij}>0]$, so that their weights
depend on the location of $(\mathbf{z}_{i},\mathbf{x}_{j})$ relative to
$(\mathbf{z},\mathbf{x)}$. This means that the closer $(\mathbf{z}%
_{i},\mathbf{x}_{j})$ is to $(\mathbf{z},\mathbf{x)}$, the greater weight is
assigned to the indicator function. Since all pairs of indices from the set
$\mathcal{R}^{+}$ satisfy the condition $\Delta_{ij}>0$, then the indicator
functions are equal to 1 for all $(i,j)\in \mathcal{R}^{+}$ as it is shown in
the expression for $p^{+}(\mathbf{z},\mathbf{x})$. At the same time, the sum
of attention weights for all pairs of indices $(r,s)\in \mathcal{R}^{+}%
\cup \mathcal{R}^{-}$ is equal to 1.

According to Nadaraya-Watson kernel regression
\cite{Nadaraya-1964,Watson-1964}, weights can be defined by means of the
kernel $K$ as a function of the distance between the vectors $\mathbf{x}_{i}$
and $\mathbf{x}$. The kernel estimates how $\mathbf{x}_{i}$ is close to
$\mathbf{x}$. Then the weight is written as follows:
\begin{equation}
a((\mathbf{z},\mathbf{x)},(\mathbf{z}_{i},\mathbf{x}_{j}))=\frac
{K((\mathbf{z},\mathbf{x)},(\mathbf{z}_{i},\mathbf{x}_{j}))}{\sum
_{(r,s)\in \mathcal{R}^{+}\cup \mathcal{R}^{-}}K((\mathbf{z},\mathbf{x)}%
,(\mathbf{z}_{r},\mathbf{x}_{s}))}.
\end{equation}

If the kernel $K((\mathbf{z},\mathbf{x)},(\mathbf{z}_{i},\mathbf{x}_{j}))$ is
Gaussian, then the attention weight is expressed through the softmax
operation:
\begin{equation}
a((\mathbf{z},\mathbf{x)},(\mathbf{z}_{i},\mathbf{x}_{j}%
))=\text{\textrm{softmax}}\left(  -\frac{\left \Vert (\mathbf{z},\mathbf{x)}%
-(\mathbf{z}_{i},\mathbf{x}_{j})\right \Vert ^{2}}{\tau}\right)  ,
\end{equation}
where $\tau$ is a tuning (temperature) parameter.

The numerator in $a((\mathbf{z},\mathbf{x)},(\mathbf{z}_{i},\mathbf{x}_{j}))$
measures the similarity between the query point $(\mathbf{z},\mathbf{x})$ and
the training pair $(\mathbf{z}_{i},\mathbf{x}_{j})$. The denominator sums over
all pairs involved in the dataset, acting as a local normalizing constant. In
a special case, when the kernels are Gaussian and concatenated vectors do not
include cross-terms between $\mathbf{z}$ and $\mathbf{x}$ that prevent
factorization, the attention weights are defined as:
\begin{equation}
a(\mathbf{z},\mathbf{x},\mathbf{z}_{i},\mathbf{x}_{j})=\frac{K(\mathbf{z}%
,\mathbf{z}_{i})\cdot K(\mathbf{x},\mathbf{x}_{j})}{\sum_{(r,s)\in
\mathcal{R}^{+}\cup \mathcal{R}^{-}}K(\mathbf{z},\mathbf{z}_{r})\cdot
K(\mathbf{x},\mathbf{x}_{s})}.
\end{equation}

Following the attention mechanism framework
\cite{Luong-etal-2015,Vaswani-etal-17}, we treat $(\mathbf{z},\mathbf{x})$ as
the \emph{query}, $(\mathbf{z}_{i},\mathbf{x}_{j})$ as \emph{keys}, and $1$ as
\emph{values}. Let us define the :
\begin{align}
\mathbf{q}  &  =\mathbf{W}_{Q}\left(  \mathbf{z},\mathbf{x}\right)  ^{\top}%
\in \mathbb{R}^{d},\nonumber \\
\mathbf{k}_{ij}  &  =\mathbf{W}_{K}\left(  \mathbf{z}_{i},\mathbf{x}%
_{j}\right)  ^{\top}\in \mathbb{R}^{d},\nonumber \\
v_{ij}  &  =\mathbf{1}[\Delta_{ij}>0],
\end{align}
where $\mathbf{W}_{Q}\in \mathbb{R}^{d\times(2d+1)}$ and $\mathbf{W}_{K}%
\in \mathbb{R}^{d\times(2d+1)}$ are learnable weight matrices. The attention
weights become:
\begin{equation}
a(\mathbf{z},\mathbf{x},\mathbf{z}_{i},\mathbf{x}_{j})=\frac{\exp \left(
\frac{1}{\sqrt{2d+1}}\mathbf{q}^{\top}\mathbf{k}_{ij}\right)  }{\sum
_{(s,r)\in \mathcal{R}^{+}\cup \mathcal{R}^{-}}\exp \left(  \frac{1}{\sqrt{2d+1}%
}\mathbf{q}^{\top}\mathbf{k}_{sr}\right)  }.
\end{equation}

Let $\mathbf{K}=\left[  \mathbf{k}_{ij}\right]  $ contain all keys and
$\mathbf{V}=\left[  v_{ij}\right]  $ contain all values. The distribution can
be expressed in the matrix form:
\begin{equation}
p^{+}(\mathbf{z},\mathbf{x})=\text{softmax}\left(  \frac{1}{\sqrt{2d+1}%
}\mathbf{q}\mathbf{K}^{\top}\right)  \mathbf{V}.
\end{equation}

When features have different importance, we introduce a diagonal weight matrix
$\mathbf{W}\in \mathbb{R}^{d\times(2d+1)}$:%
\begin{equation}
\mathbf{W}=\operatorname{diag}(\omega_{1},\omega_{2},\dots,\omega_{2d+1}),
\end{equation}
where $\omega_{i}\geq0$ represents the importance of the $i$-th feature
dimension. Hence, the weighted queries and weighted keys are $\widetilde
{\mathbf{Q}}=\mathbf{Q}\mathbf{W}^{1/2}$ and $\widetilde{\mathbf{K}%
}=\mathbf{K}\mathbf{W}^{1/2}$, respectively, where $\mathbf{W}^{1/2}%
=\operatorname{diag}(\sqrt{\omega_{1}},\dots,\sqrt{\omega_{2d+1}})$.

The model learns matrices $\mathbf{W}_{Q}$ and $\mathbf{W}_{K}$ by minimizing
the log-likelihood loss with the $L_{2}$-regularization:
\begin{align}
\mathcal{L}(\mathbf{p} &  \mid \mathbf{W}_{Q},\mathbf{W}_{K})=-\sum
_{(i,j)\in \mathcal{R}^{+}}\log \left(  p^{+}(\mathbf{z}_{i},\mathbf{x}%
_{j})\right)  -\sum_{(i,j)\in \mathcal{R}^{-}}\log \left(  p^{-}(\mathbf{z}%
_{i},\mathbf{x}_{j})\right)  \nonumber \\
&  +\eta \left(  \Vert \mathbf{W}_{Q}\Vert^{2}+\Vert \mathbf{W}_{K}\Vert
^{2}\right)  ,
\end{align}
where $\eta$ is the hyperparameter controlling the strength of the $L_{2}%
$-regularization. 

During inference, the probability $p^{+}(\mathbf{x},\mathbf{x})$ is computed
for a new instance $\mathbf{x}$ under conditions $A=1,...,m$. Suppose that
$\gamma$ is the threshold of the probability $p^{+}(\mathbf{x},\mathbf{x})$
for decision making about the positive treatment effect. Then the number of
doses is selected minimizing the difference between $p^{+}(\mathbf{x}%
,\mathbf{x})$ and $\gamma$ by $p^{+}(\mathbf{x},\mathbf{x})\geq \gamma$.

The quantity $p^{+}(\mathbf{z},\mathbf{x})$ can be viewed as a \emph{causal
risk} function reflecting the probability that the treatment yields a
beneficial outcome for an individual with covariates $\mathbf{x}$. Direct
estimation of $\Delta$ at the individual level is infeasible because, for each
individual, only one potential outcome (either under treatment or control) is
observed. The proposed approach employs a \emph{contrastive} approach inspired
by classification: by forming pairs $(\mathbf{z}_{i},h_{i})$ and
$(\mathbf{x}_{j},y_{j})$ from treatment and control groups, respectively, we
approximate the distribution of individual effects indirectly. The sign of the
observed difference $h_{i}-y_{j}$ for these pairs acts as a surrogate
indicator of the unobserved individual effect, assuming that individuals with
similar covariates have similar potential outcomes. This assumption aligns
with the \emph{conditional exchangeability} condition common in causal inference.

By aggregating large numbers of such pairs and weighting their contributions
according to covariate similarity (via attention or kernel methods), the model
effectively estimates the probability that a new individual with covariates
$\mathbf{x}$ would benefit from treatment instances $\mathbf{z}$. The
likelihood that $\Delta>0$ for a given pair is thus approximated by the label
indicating positive outcomes of their difference, up to the approximation
quality provided by the weighting scheme.

\section{Numerical Experiments}

To assess model performance, we divide all experiments into two parts. The
first part of numerical experiments can be conducted on both synthetic and
real data. Predicted probabilities $\Pr \left \{  {\Delta>0\mid \mathbf{Z}%
=\mathbf{z}_{i},\mathbf{X}=\mathbf{x}_{j}}\right \}  $ are calculated for each
pair $(\mathbf{z}_{i},\mathbf{x}_{j})$ from the testing set. In this case, we
do not need to consider only pairs $(\mathbf{x},\mathbf{x})$ because
Dose-AIPTB allows us to estimate $p^{+}(\mathbf{z},\mathbf{x})$ for arbitrary
pairs composed from the treatment and control groups. The corresponding
validation metrics will be called Val 1. The second part is implemented using
only synthetic datasets. It estimates IPTB for patients whose feature vectors
are identical in both the treatment and control groups. The corresponding
outcomes $h_{i}$ and $y_{j}$ are generated in accordance with the dataset
rules. After generating control and treatment outcomes for many feature
vectors, for each pair $(\mathbf{x},\mathbf{x})$, we can calculate the
predicted probability $p^{+}(\mathbf{x},\mathbf{x})=\Pr \left \{  {\Delta
>0\mid \mathbf{Z}=\mathbf{x},\mathbf{X}=\mathbf{x}}\right \}  $ using the
proposed model. The corresponding validation metrics will be called Val 2.

For quantitative evaluation, we provide the area under the receiver operating
characteristic curve (ROC-AUC). Each metric value is obtained by averaging
results from five-fold stratified cross-validation repeated ten times with
different random seeds to ensure statistical reliability. For every dataset,
we provide two types of graphs with the ROC-curves. The first type (left
graphs) illustrates the training, validation-1 (Val 1) and validation-2 (Val
2) ROC-curves for Dose-AIPTB, where Val 1 and Val 2 correspond to validation
on pairs $(\mathbf{z},\mathbf{x})$ and $(\mathbf{x},\mathbf{x})$, respectively.

Dose-AIPTB is compared with meta-learners: the T-learner, S-learner,
X-learner, and DR-learner. Each meta-learner is trained using the random
forest regressor consisting of 200 random trees. It should be noted that the
meta-learners return CATE values, so the sign of the CATE is converted into
the IPTB class label. At that, meta-learners are trained on every dose.

\subsection{Synthetic data}

We study the proposed model Dose-AIPTB on synthetic datasets: \emph{Simple,
Linear, Step-wise, Spiral, Power, Weibull}. They are generated in accordance
with the following functions of the same name.

\begin{enumerate}
\item \emph{Simple}: Instances are generated such that
\begin{equation}
y(A)=(1-0.25\cdot A)\cdot x^{(1)}+(0.25+0.25\cdot A)\cdot x^{(2)}.
\end{equation}

Here $A$ is uniformly generated from $\{1,2,3\}$ for treatments and $A=0$ for
controls. Numbers of controls and treatments are $c=350$ and $t=50$, respectively.

\item \emph{Linear}: Feature values $x^{(i)}$ are uniformly generated from
$[0,1]$, i.e., $x^{(i)}\sim \mathcal{U}(0,1)$, values of $y$ are generated
using the linear function:%
\begin{equation}
y(A)=2\cdot(1-0.2\cdot A)\cdot x^{(1)}+4\cdot x^{(2)}+8\cdot0.2\cdot A\cdot
x^{(3)}.
\end{equation}

Numbers of controls and treatments are the same as in the Simple dataset;
$A\in \{0,1,2,3,4,5\}$.

\item \emph{Step-wise}: Instances are generated such that
\begin{align}
y(A)  &  =(1-0.25\cdot A\cdot \mathbb{I}\left(  x^{(1)}<0.5\right)  )\cdot
x^{(1)}\nonumber \\
&  +(0.25+0.25\cdot A)\cdot x^{(2)}\cdot \mathbb{I}\left(  x^{(2)}<0.5\right)
.
\end{align}

Here $A\in \{0,1,2,3\}$. The dosage increases the weight of $x^{(2)}$ in the
outcome while decreasing the weight of $x^{(1)}$. The treatment effect in this
dataset can be either positive or negative. Parameters of the Step-wise
dataset are the same as for the Simple dataset, but when $x^{(1)}>0.5$, its
coefficient becomes $1$, and when $x^{(2)}>0.5$, its coefficient is set to
$0$. This rule for $x^{(1)}>0.5$ removes the negative effect of the dosage,
while for $x^{(2)}>0.5$, it removes the positive effect. When both
$x^{(1)}>0.5$ and $x^{(2)}>0.5$, the dosage becomes irrelevant ($y$ depends
linearly only on $x^{(1)}$).

\item \emph{Spiral}: Vectors $\mathbf{x}\in \mathbb{R}^{5}$ are generated by
using the Archimedean spiral as follows:
\begin{equation}
\mathbf{x}=(t\sin(t),t\cos(t),...,t\sin(t\cdot d/2),t\cos(t\cdot d/2)),
\end{equation}
for even $d$, and
\begin{equation}
\mathbf{x}=(t\sin(t),t\cos(t),...,t\sin(t\cdot \left \lceil d/2\right \rceil )).
\end{equation}
for odd $d$. Values of $y$ are generated as
\begin{equation}
y(A)=(a+0.2\cdot A)\cdot t-1.5\cdot A\cdot b,
\end{equation}
where $a\sim \mathcal{U}(0.6,1)$, $b\sim \mathcal{U}(0.6,1)$, $t\sim
\mathcal{U}(1,12)$, $d=5$, $A$ is uniformly generated from $\{1,2,3,4,5\}$. If
$t$ is large, increasing $A$ will increase $y(A)$, otherwise, it will decrease
$y(A)$. Here $c=400$ and $t=200$.

\item \emph{Power}:\emph{ }Feature vectors $\mathbf{x}\in \mathbb{R}^{5}$ are
generated by using the following representation:
\begin{equation}
\mathbf{x}=(t^{1/\sqrt{d}},t^{2/\sqrt{d}},...,t^{d/\sqrt{d}}).
\end{equation}
Outcomes are computed as:%
\begin{equation}
y(A)=(a-0.2\cdot A)\cdot \exp \left(  -\frac{(t-s)^{2}}{b-0.1\cdot(A-5)}\right)
.
\end{equation}

Here $a\sim \mathcal{U}(9,10)$, $b\sim \mathcal{U}(0.5,1)$, $t\sim
\mathcal{U}(0,7)$, $s=3.5$, $d=5$, $A$ is uniformly generated from
$\{1,2,3,4,5\}$. The effect of $A$ on $y(A)$ is positive when $(t-s)$ is large
and negative when $(t-s)$ is small.

\item \emph{Weibull}: Each observation has two features, $x^{(1)}$ and
$x^{(2)}$ sampled uniformly from $0$ to $1$. The event time (outcome) is
generated from the Weibull distribution with shape parameter $k=3.0$ as
follows:%
\begin{equation}
T=\frac{y(A)}{\Gamma \left(  1+\frac{1}{k}\right)  }\cdot \left(  -\log
(u)\right)  ^{\frac{1}{k}}.
\end{equation}
Here $\Gamma(\cdot)$ is the gamma function; $u\sim \mathcal{U}(0,1)$; $y(A)$ is
determined as in the Simple dataset; $A\in \{0,1,2,3\}$.
\end{enumerate}

\begin{figure}
[ptb]
\begin{center}
\includegraphics[
height=1.9in,
width=4.8in
]%
{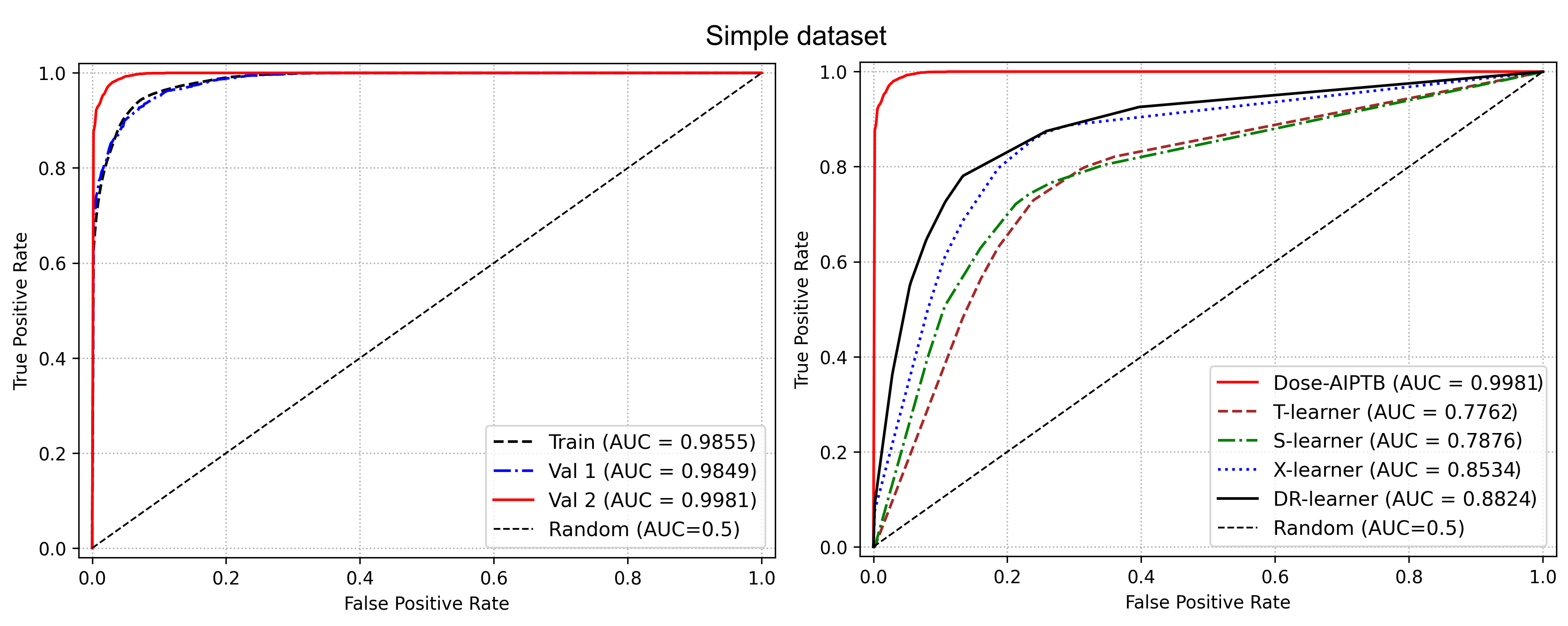}%
\caption{Left plot: The ROC curves and ROC-AUC scores obtained on the
training, Val 1, and Val 2 sets for Dose-AIPTB trained on the Simple dataset.
Right plot: Comparison of ROC curves and ROC-AUC scores for Dose-AIPTB and the
meta-learners.}%
\label{f:roc_curves_aggregated_simple}%
\end{center}
\end{figure}

Fig. \ref{f:roc_curves_aggregated_simple} displays two side-by-side Receiver
Operating Characteristic (ROC) plots obtained for the \emph{Simple} dataset.
The left plot shows the performance of the proposed model Dose-AIPTB on
training and validation sets, where all curves (Train, Val 1, Val 2) are
tightly clustered near the top left with high AUC scores ranging from roughly
$0.985$ to $0.998$. The right plot compares different meta-learner algorithms,
showing that the Dose-AIPTB model (red solid line) achieves the best
performance with an AUC of $0.9981$, significantly outperforming the
T-learner, S-learner, X-learner, and DR-learner.%

\begin{figure}
[ptb]
\begin{center}
\includegraphics[
height=2.1in,
width=2.9in
]%
{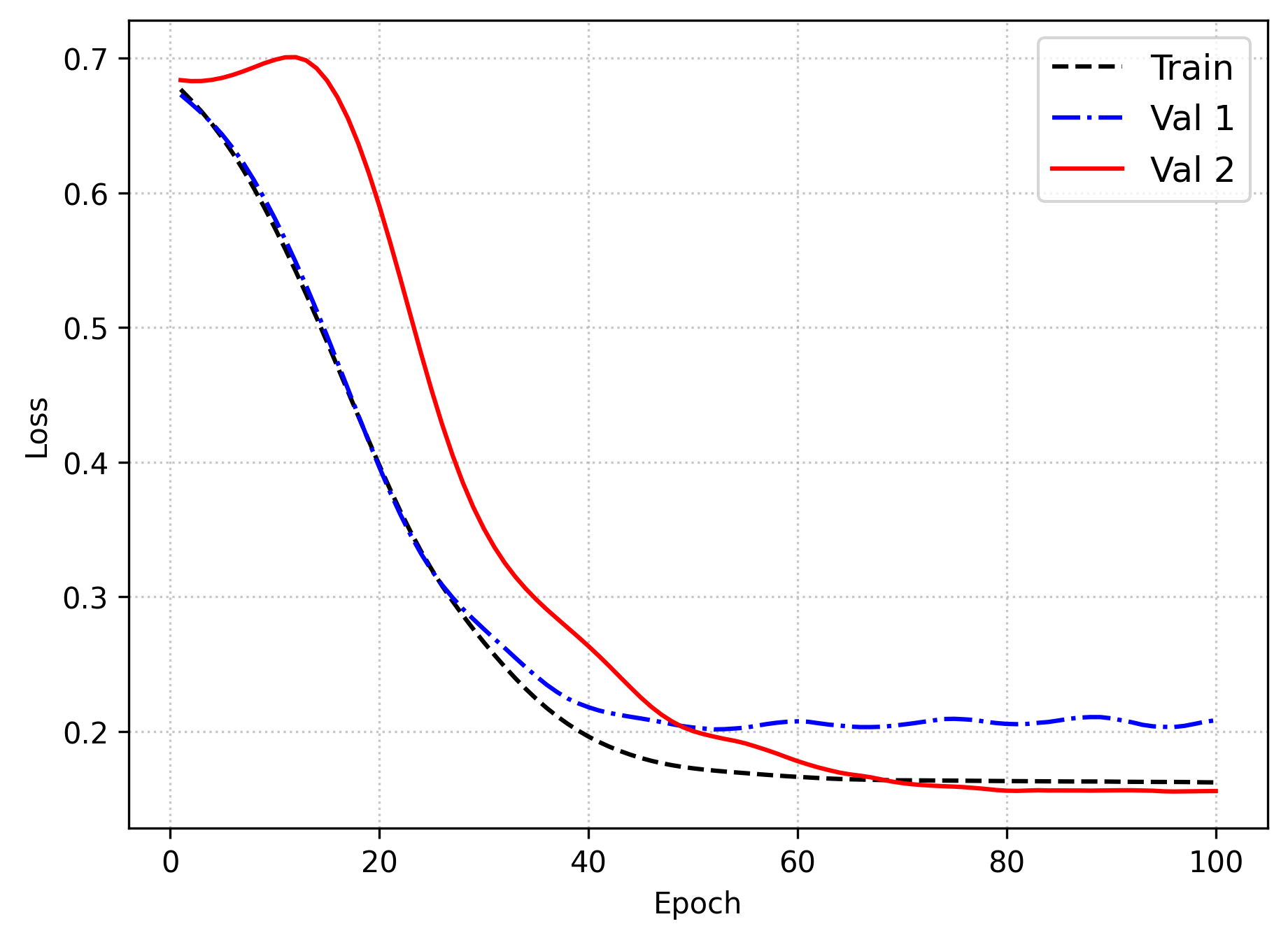}%
\caption{Training and validation loss functions for the Simple dataset}%
\label{f:loss_aggregated_simple}%
\end{center}
\end{figure}

Fig. \ref{f:loss_aggregated_simple} illustrates values of training (Train) and
two testing (Val 1 and Val 2) loss functions depending on the epoch numbers
for the Simple dataset. One can see from the figure that two curves (Train and
Val 1) follow an almost identical path. They drop rapidly until about epoch
40. Afterward, they flatten out, slowly decreasing to a final loss of around
$0.1$. The third curve (Val 2) behaves differently. It is decreasing to a
final loss of around $0.2$.%

\begin{figure}
[ptb]
\begin{center}
\includegraphics[
height=1.9in,
width=4.8in
]%
{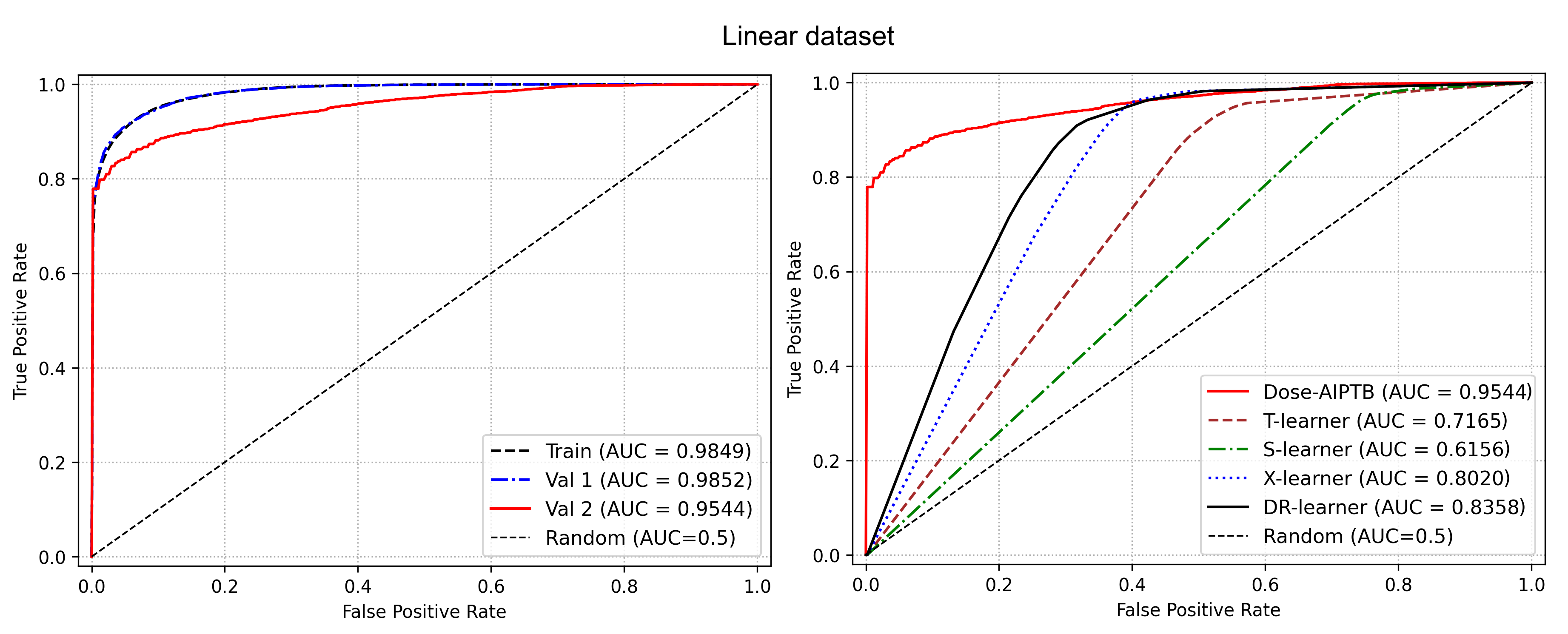}%
\caption{Left plot: The ROC curves and ROC-AUC scores obtained on the
training, Val 1, and Val 2 sets for Dose-AIPTB trained on the Linear dataset.
Right plot: Comparison of ROC curves and ROC-AUC scores for Dose-AIPTB and the
meta-learners.}%
\label{f:roc_curves_aggregated_linear}%
\end{center}
\end{figure}

Fig. \ref{f:roc_curves_aggregated_linear} displays similar ROC plots obtained
for the \emph{Linear} dataset. The left plot shows the performance of the
proposed model Dose-AIPTB on training and validation sets. The right plot also
demonstrates that the Dose-AIPTB model (red solid line) achieves the best
performance with an AUC of $0.9544$, significantly outperforming the
T-learner, S-learner, X-learner, and DR-learner.%

\begin{figure}
[ptb]
\begin{center}
\includegraphics[
height=2.1in,
width=2.9in
]%
{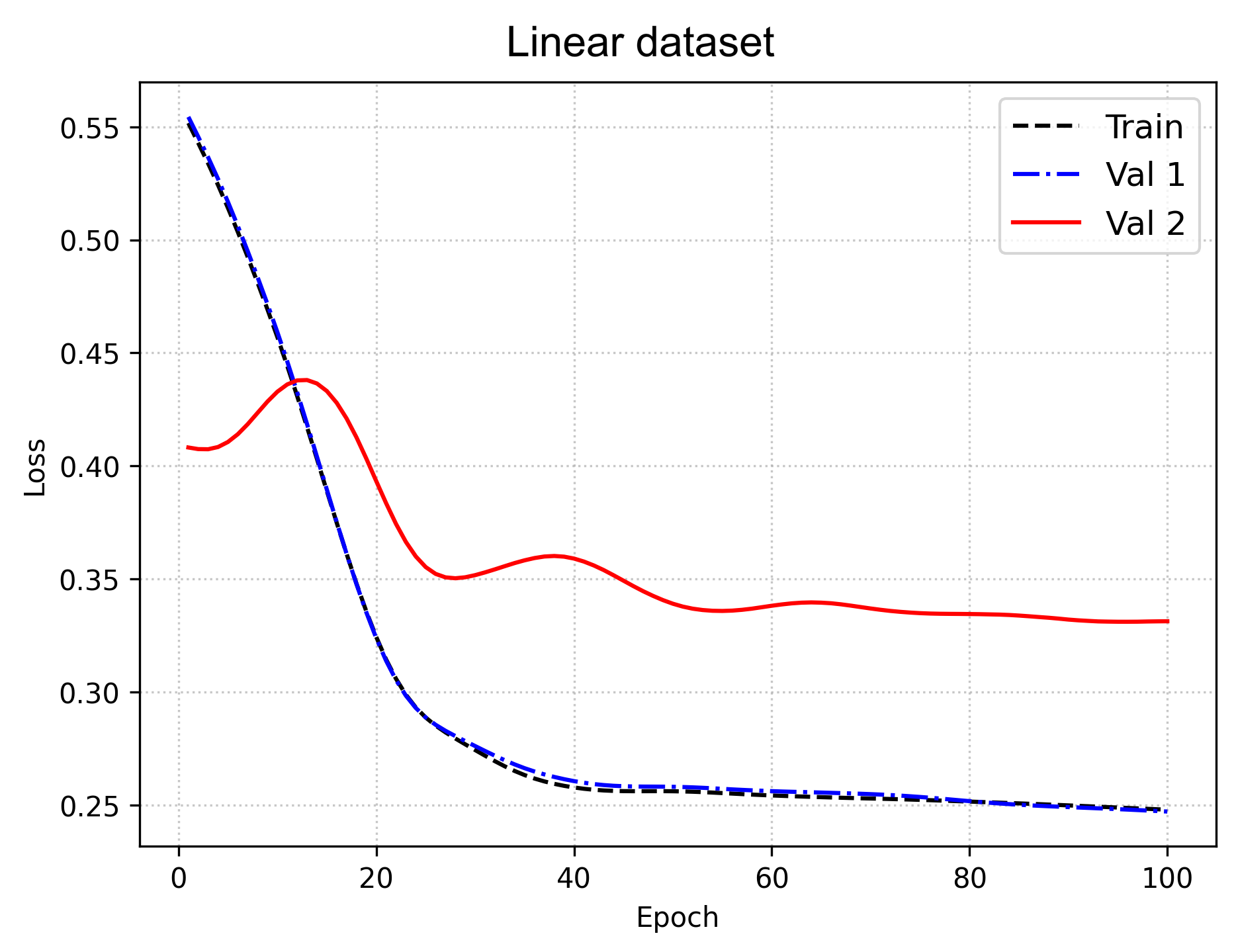}%
\caption{Training and validation loss functions for the Linear dataset}%
\label{f:loss_aggregated_linear}%
\end{center}
\end{figure}

Fig. \ref{f:loss_aggregated_linear} illustrates values of training (Train) and
two testing (Val 1 and Val 2) loss functions depending on the epoch numbers
for the Linear dataset. Two curves (Train and Val 1) also follow an almost
identical path. The third curve (Val 2) starts lower than the others and
plateaus at a significantly higher loss level compared to the Train and Val 1
functions suggesting the model performs worse on this specific validation set.%

\begin{figure}
[ptb]
\begin{center}
\includegraphics[
height=1.9in,
width=4.8in
]%
{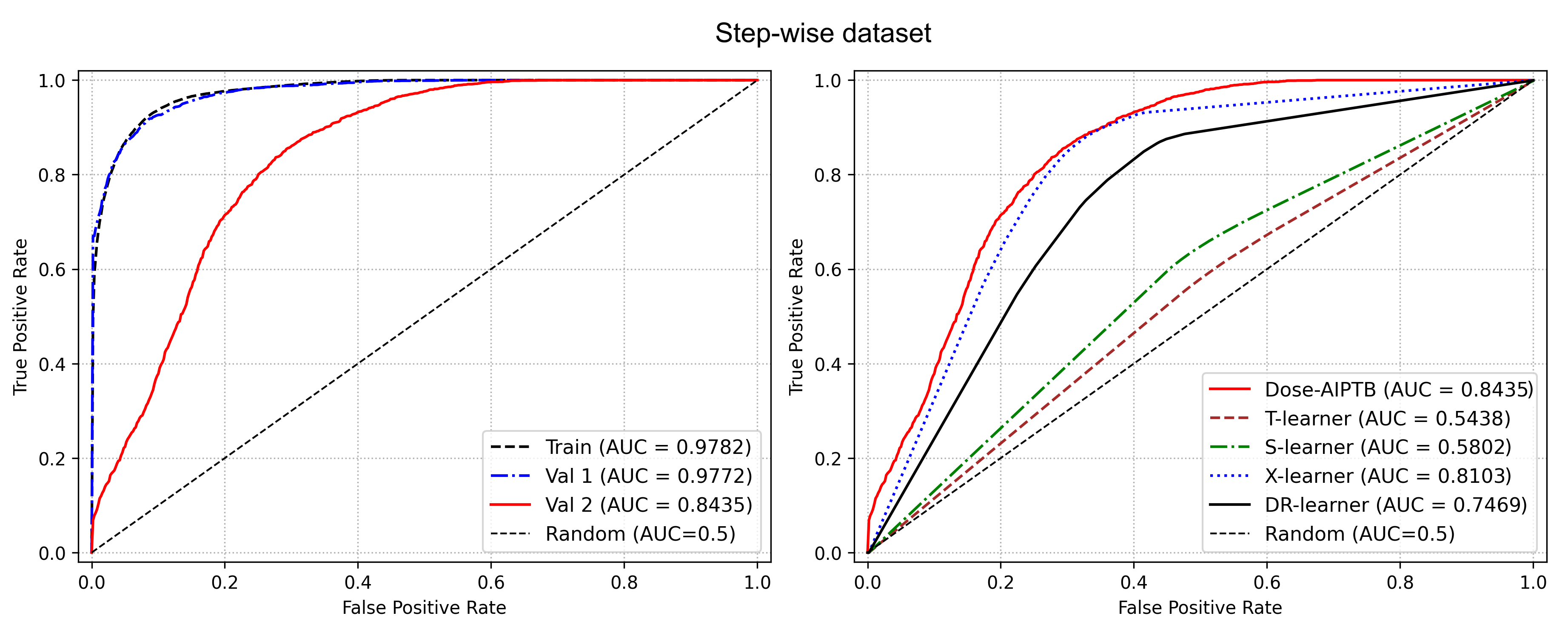}%
\caption{Left plot: The ROC curves and ROC-AUC scores obtained on the
training, Val 1, and Val 2 sets for Dose-AIPTB trained on the Step-wise
dataset consisting of 350 controls and 50 treatments. Right plot: Comparison
of ROC curves and ROC-AUC scores for Dose-AIPTB and the meta-learners.}%
\label{f:roc_curves_aggregated_step_350c_50t}%
\end{center}
\end{figure}

\begin{figure}
[ptb]
\begin{center}
\includegraphics[
height=1.9in,
width=4.8in
]%
{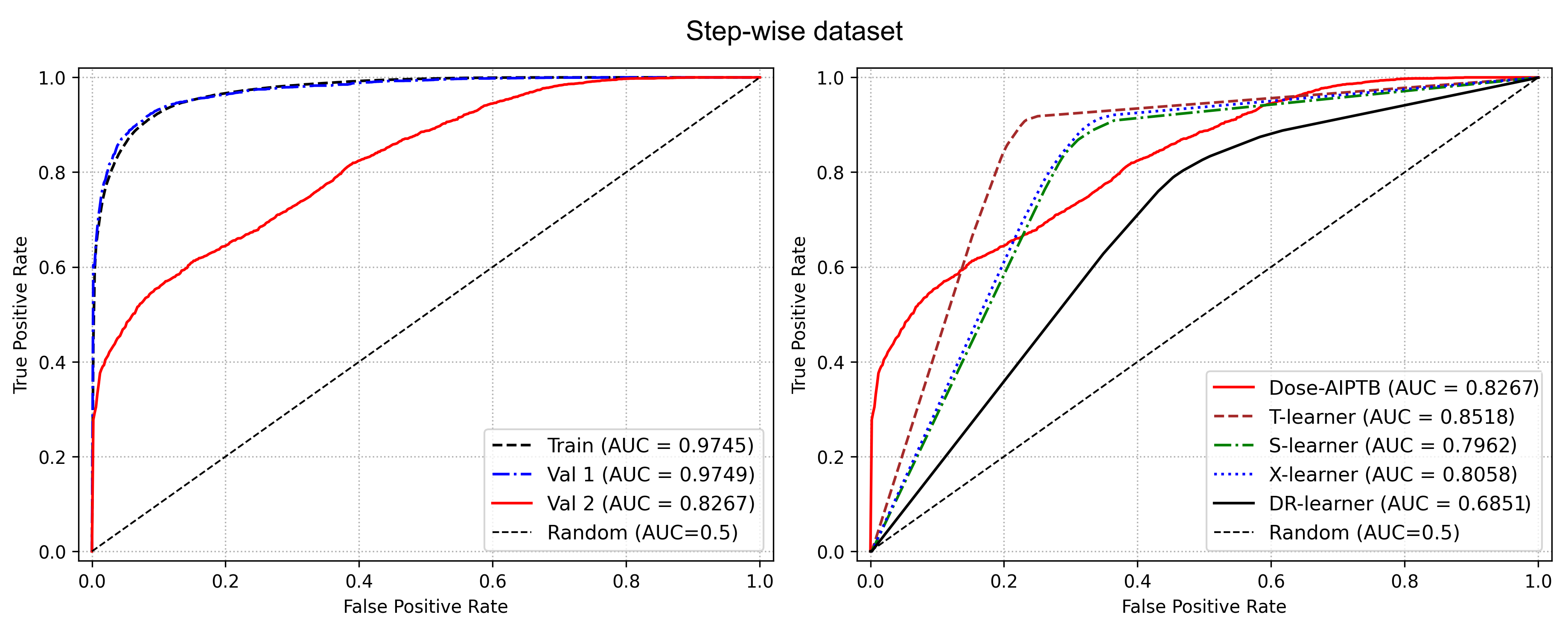}%
\caption{ROC-AUC scores obtained on the training, Val 1, and Val 2 sets for
Dose-AIPTB trained on the Step-wise dataset consisting of 50 controls and 350
treatments. Right plot: Comparison of ROC curves and ROC-AUC scores for
Dose-AIPTB and the meta-learners.}%
\label{f:roc_curves_aggregated_step_50c_350}%
\end{center}
\end{figure}

A more complex dataset is \emph{Step-wise}. Similar results for this dataset
under condition of training on 350 controls and 50 treatments are shown in
Fig. \ref{f:roc_curves_aggregated_step_350c_50t}. It can be seen from the
figure that Dose-AIPTB is comparable with the X-learner. It is interesting to
point out that results change if the models are trained on 50 controls and 350
treatments as shown in Fig. \ref{f:roc_curves_aggregated_step_50c_350}.%

%

\begin{figure}
[ptb]
\begin{center}
\includegraphics[
height=1.9in,
width=4.8in
]%
{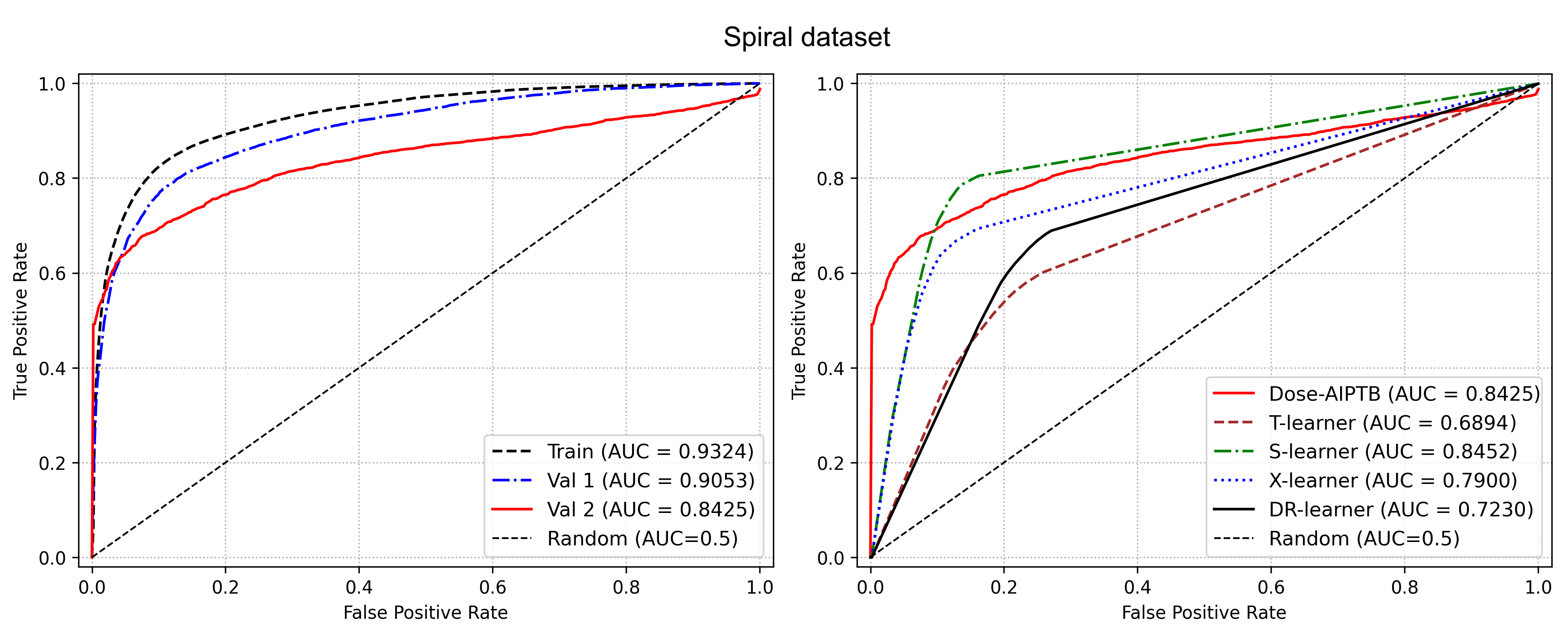}%
\caption{Left plot: The ROC curves and ROC-AUC scores obtained on the
training, Val 1, and Val 2 sets for Dose-AIPTB trained on the Spiral dataset.
Right plot: Comparison of ROC curves and ROC-AUC scores for Dose-AIPTB and the
meta-learners.}%
\label{f:roc_curves_aggregated_spiral}%
\end{center}
\end{figure}

\begin{figure}
[ptb]
\begin{center}
\includegraphics[
height=1.9in,
width=4.8in
]%
{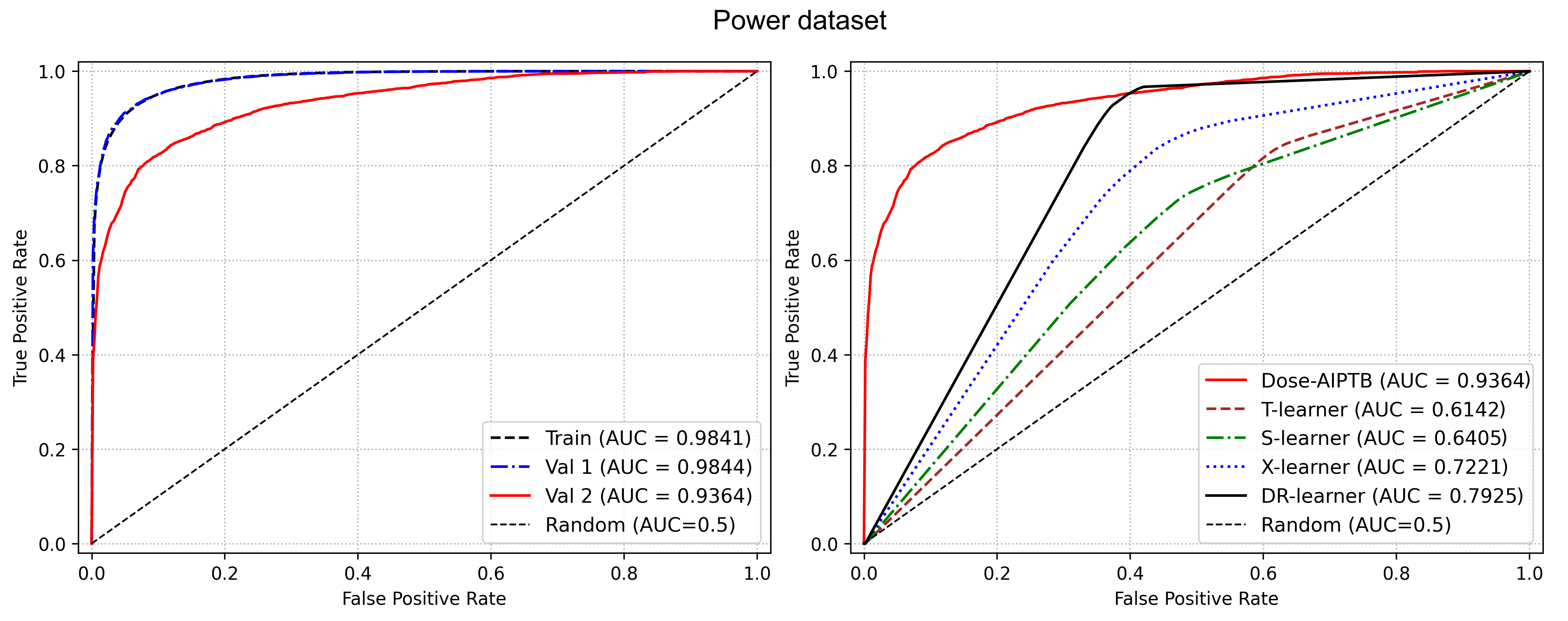}%
\caption{Left plot: The ROC curves and ROC-AUC scores obtained on the
training, Val 1, and Val 2 sets for Dose-AIPTB trained on the Power dataset.
Right plot: Comparison of ROC curves and ROC-AUC scores for Dose-AIPTB and the
meta-learners.}%
\label{f:roc_curves_aggregated_power}%
\end{center}
\end{figure}

Figs. \ref{f:roc_curves_aggregated_spiral} and
\ref{f:roc_curves_aggregated_power} illustrate results for the \emph{Spiral}
and \emph{Power} datasets, respectively.%

%

\begin{figure}
[ptb]
\begin{center}
\includegraphics[
height=5.7in,
width=4.8in
]%
{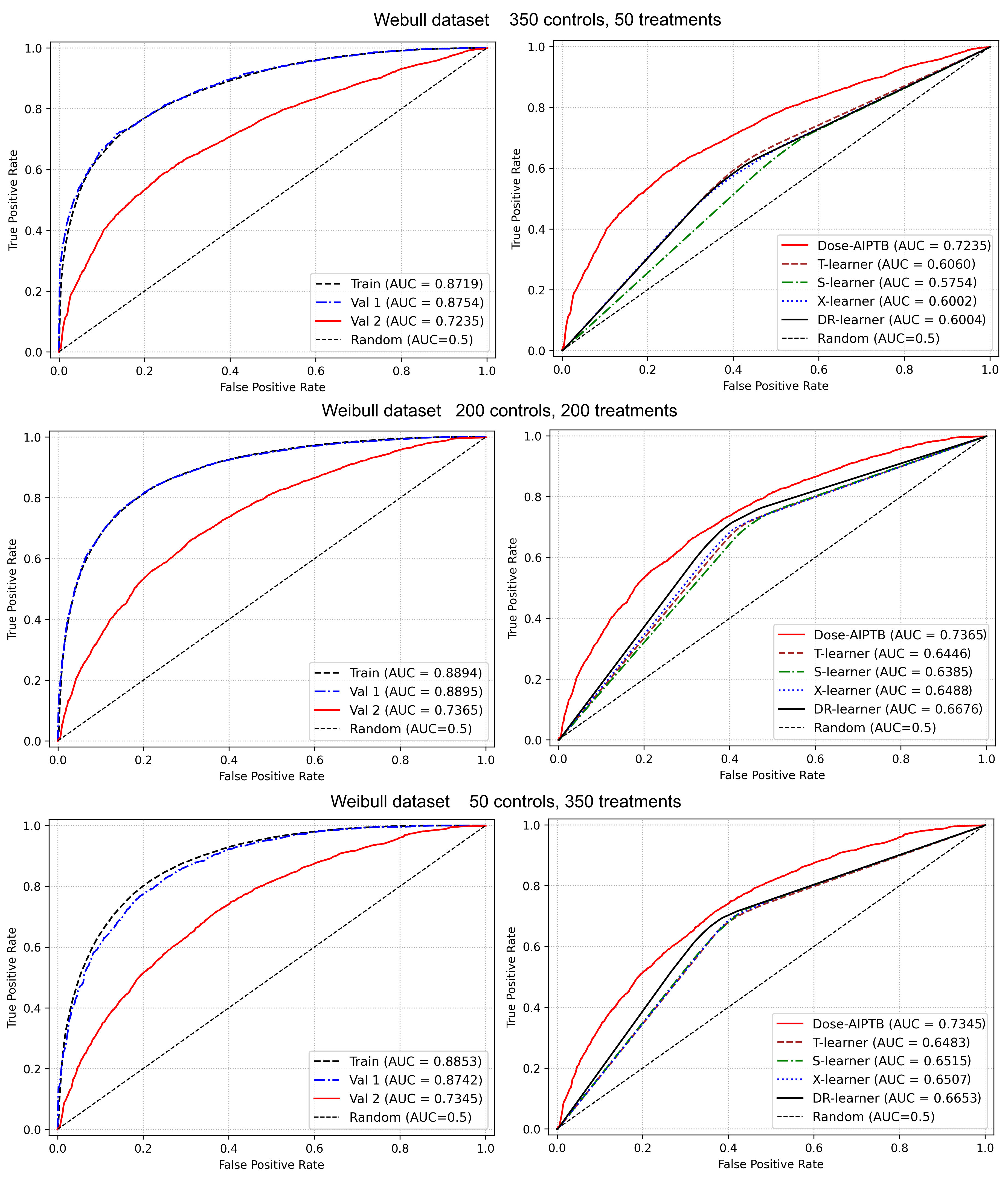}%
\caption{Left plots: The ROC curves and ROC-AUC scores obtained on the
training, Val 1, and Val 2 sets for Dose-AIPTB trained on the Weibull datasets
which differ by numbers of controls and treatments. Right plots: The
corresponding comparison of ROC curves and ROC-AUC scores for Dose-AIPTB and
the meta-learners.}%
\label{f:roc_curves_aggregated_weibul}%
\end{center}
\end{figure}

Fig. \ref{f:roc_curves_aggregated_weibul} shows results for the \emph{Weibull}
datasets generated with different numbers of controls and treatments. It can
be seen from the plots that Dose-AIPTB does not change its performance in
contrast to other models. The extended results are shown in Table
\ref{t:Weibull_1}. This table presents AUC values with standard deviations for
five models trained on the Weibull dataset under varying ratios of controls
(c) to treatments (t). The Dose-AIPTB method consistently achieves the highest
performance across all seven experimental settings, maintaining AUC scores
between $0.724$ and $0.737$ regardless of the sample split. In contrast, the
meta-learner baselines (T-learner, S-learner, X-learner, and DR-learner)
exhibit lower and more variable performance, with AUC values generally ranging
from $0.575$ to $0.675$. Notably, the performance of Dose-AIPTB remains
remarkably stable even when the treatment group becomes very small (e.g., 50
treatments vs. 350 controls), whereas other models tend to degrade more
significantly under imbalanced conditions. The most challenging configuration
appears to be the extreme imbalance of 350 controls to 50 treatments, where
all models except Dose-AIPTB drop to their lowest or near-lowest scores.%

%

\begin{table}[tbp] \centering
\caption{AUC values for the models trained on the Weibull dataset by different numbers of controls (c) and treatments (t)}%
\begin{tabular}
[c]{cccccc}\hline
$c/t$ & Dose-AIPTB & T-learner & S-learner & X-learner & DR-learner\\ \hline
$50/350$ & $0.734\pm0.020$ & $0.648\pm0.010$ & $0.651\pm0.011$ &
$0.651\pm0.012$ & $0.665\pm0.019$\\ \hline
$100/300$ & $0.731\pm0.019$ & $0.623\pm0.026$ & $0.619\pm0.027$ &
$0.622\pm0.018$ & $0.658\pm0.021$\\ \hline
$150/250$ & $0.737\pm0.019$ & $0.657\pm0.027$ & $0.654\pm0.024$ &
$0.656\pm0.025$ & $0.675\pm0.024$\\ \hline
$200/200$ & $0.736\pm0.020$ & $0.645\pm0.021$ & $0.639\pm0.015$ &
$0.649\pm0.016$ & $0.668\pm0.020$\\ \hline
$250/150$ & $0.734\pm0.019$ & $0.655\pm0.021$ & $0.651\pm0.018$ &
$0.655\pm0.015$ & $0.669\pm0.020$\\ \hline
$300/100$ & $0.736\pm0.018$ & $0.620\pm0.012$ & $0.605\pm0.014$ &
$0.624\pm0.017$ & $0.666\pm0.019$\\ \hline
$350/50$ & $0.724\pm0.021$ & $0.606\pm0.016$ & $0.575\pm0.024$ &
$0.600\pm0.021$ & $0.600\pm0.037$\\ \hline
\end{tabular}
\label{t:Weibull_1}%
\end{table}%

\subsection{Real data}

Our empirical evaluation leverages the widely adopted Infant Health and
Development Program (IHDP) dataset which can be viewed as a benchmark resource
frequently employed for heterogeneous treatment effect (HTE) estimation
\cite{Hill-2011}. Originally compiled to assess how specialist-conducted home
visits influence later cognitive outcomes in preterm infants, the dataset
comprises 747 participants characterized by 25 covariates: 6 continuous and 19
binary variables capturing key attributes of both infants and their mothers.
Notably, the experimental setup involves 139 distinct treatment
configurations. The dataset is publicly available via the repository at
\url{https://github.com/vdorie/npci}, facilitating reproducibility and
comparative analysis across causal inference methodologies. Due to the large
dimension of the instances in the dataset, we introduce weights of features
such that the feature with index $2d+1$ (the dose feature) has the weight
$\omega_{2d+1}=0.3$, other features have weights $\omega_{i}=0.7/(2d)$,
$i=1,...,d$.

\begin{figure}
[ptb]
\begin{center}
\includegraphics[
height=1.9in,
width=4.8in
]%
{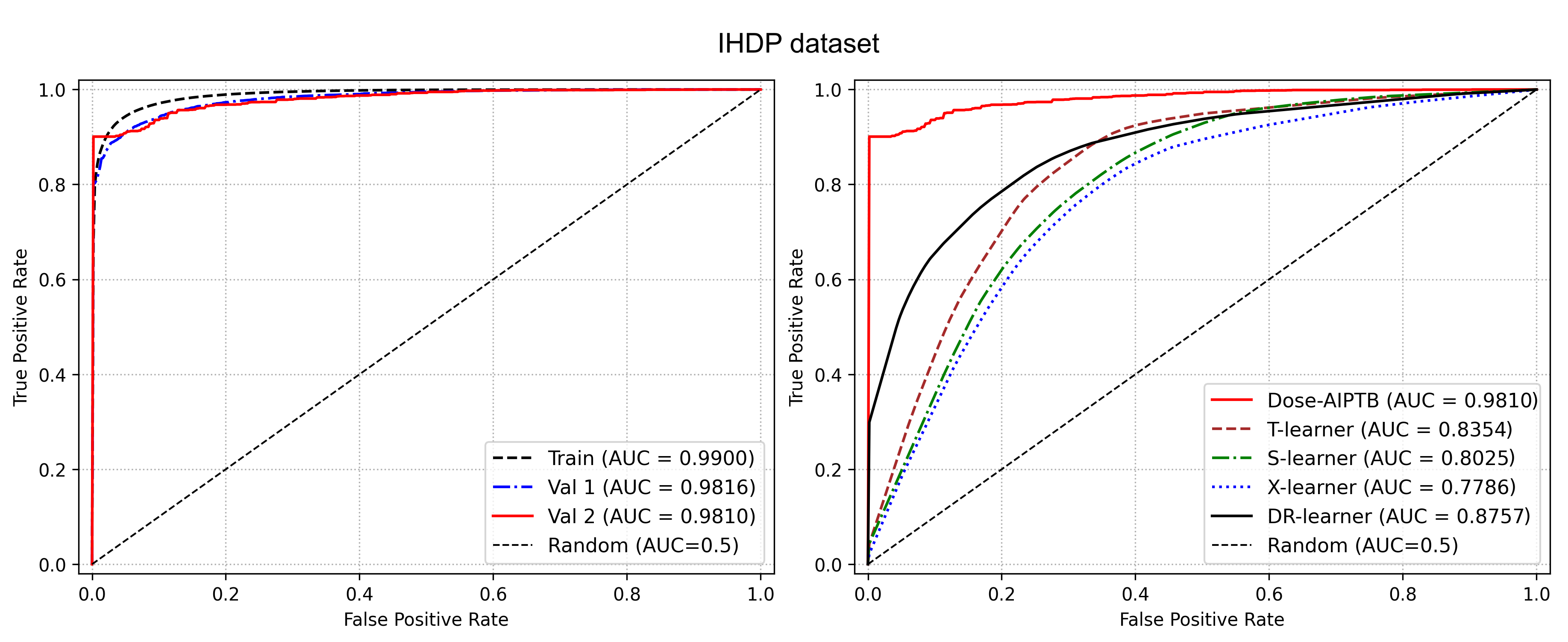}%
\caption{Left plot: The ROC curves and ROC-AUC scores obtained on the
training, Val 1, and Val 2 sets for Dose-AIPTB trained on the IHDP dataset.
Right plot: Comparison of ROC curves and ROC-AUC scores for Dose-AIPTB and the
meta-learners.}%
\label{f:roc_curves_aggregated_ihdp}%
\end{center}
\end{figure}

\begin{figure}
[ptb]
\begin{center}
\includegraphics[
height=2.1in,
width=3.0in
]%
{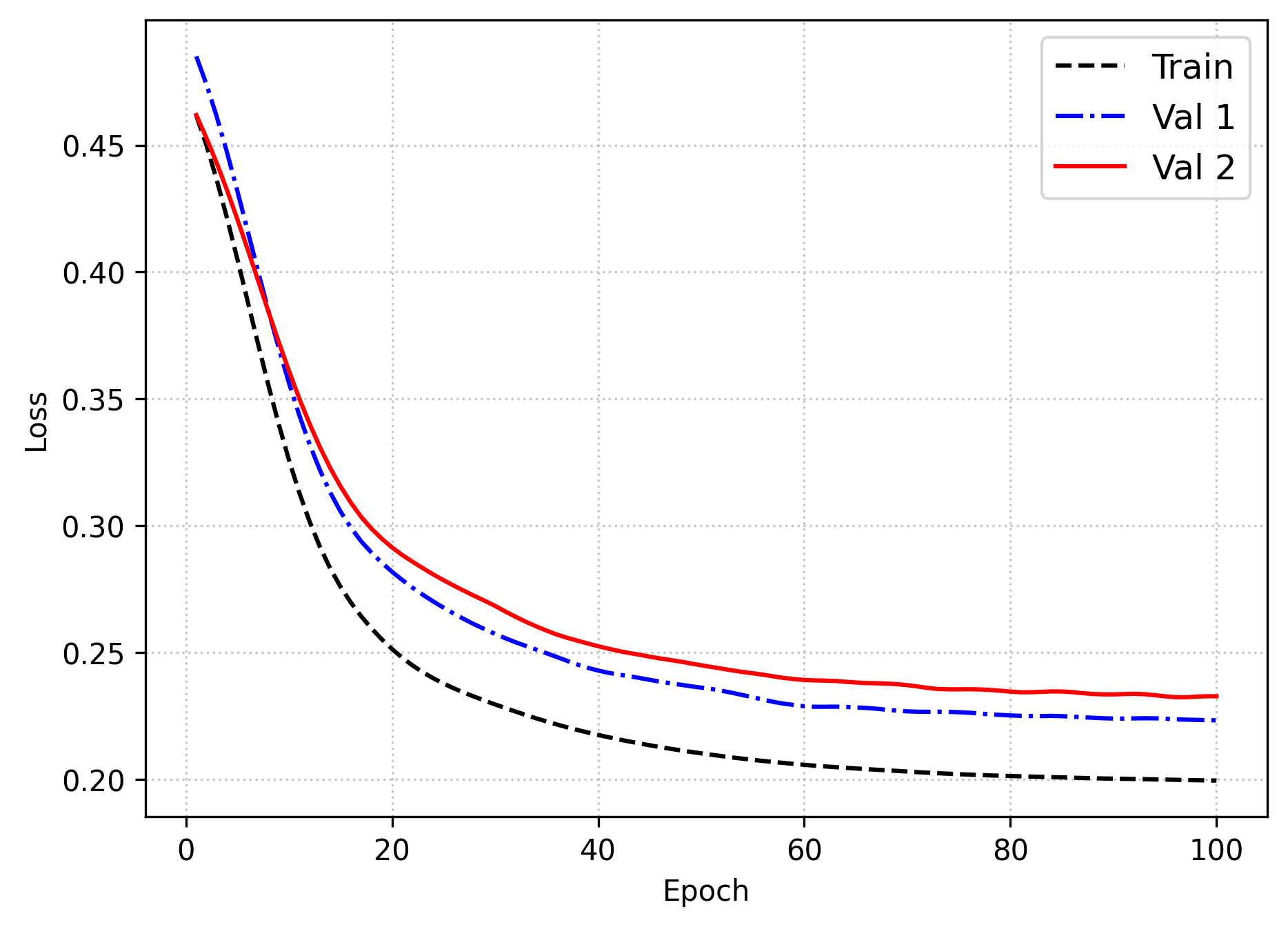}%
\caption{Training and validation loss functions for the IHDP dataset}%
\label{f:loss_aggregated_ihdp}%
\end{center}
\end{figure}

Fig. \ref{f:roc_curves_aggregated_ihdp} displays ROC plots obtained for the
\emph{IHDP} dataset. The left plot again shows the performance of the proposed
model Dose-AIPTB. The right plot compares different meta-learner algorithms,
showing that the Dose-AIPTB model achieves the best performance with an AUC of
$0.9810$, significantly outperforming the T-learner, S-learner, X-learner, and
DR-learner. Fig. \ref{f:loss_aggregated_ihdp} illustrates the loss functions
depending on the epoch numbers for the IHDP dataset. All three curves show a
steep decline in loss initially, dropping from above $0.45$ to below $0.30$
within the first 20 epochs, before the rate of improvement slows
significantly. By the end of the training, the training set reaches the lowest
loss near $0.20$, whereas the validation sets plateau at slightly higher
values, with Val 2 maintaining the highest loss of roughly $0.23$.%

Table \ref{t:all_datasets} provides AUC values for all considered datasets
obtained by Dose-AIPTB and all studied meta-models. This table presents the
AUC values, including standard deviations, for five different models trained
across seven distinct datasets. The models being compared are Dose-AIPTB,
T-learner, S-learner, X-learner, and DR-learner. Dose-AIPTB demonstrates
superior performance in the majority of the scenarios. However, on the Spiral
dataset, the S-learner slightly outperforms Dose-AIPTB with an AUC of $0.849$
compared to $0.843$. The Weibull dataset appears to be the most challenging
for all algorithms, resulting in the lowest overall AUC values which range
from roughly $0.575$ to $0.724$.%

\begin{table}[tbp] \centering
\caption{AUC values for the models trained on different datasets}%
\begin{tabular}
[c]{cccccc}\hline
Dataset & Dose-AIPTB & T-learner & S-learner & X-learner & DR-learner\\ \hline
Simple & $\mathbf{0.986\pm0.011}$ & $0.767\pm0.031$ & $0.778\pm0.036$ &
$0.831\pm0.058$ & $0.847\pm0.063$\\ \hline
Step-wise & $\mathbf{0.844\pm0.021}$ & $0.544\pm0.031$ & $0.580\pm0.022$ &
$0.811\pm0.038$ & $0.747\pm0.052$\\ \hline
Linear & $\mathbf{0.954\pm0.008}$ & $0.716\pm0.021$ & $0.616\pm0.015$ &
$0.802\pm0.028$ & $0.836\pm0.049$\\ \hline
Spiral & $0.843\pm0.041$ & $0.689\pm0.019$ & $\mathbf{0.845\pm0.015}$ &
$0.790\pm0.021$ & $0.723\pm0.017$\\ \hline
Power & $\mathbf{0.936\pm0.034}$ & $0.614\pm0.02$ & $0.640\pm0.037$ &
$0.722\pm0.044$ & $0.793\pm0.011$\\ \hline
Weibull & $\mathbf{0.724\pm0.021}$ & $0.606\pm0.016$ & $0.575\pm0.024$ &
$0.600\pm0.021$ & $0.600\pm0.037$\\ \hline
IHDP-100 & $\mathbf{0.981\pm0.018}$ & $0.835\pm0.08$ & $0.802\pm0.094$ &
$0.779\pm0.104$ & $0.876\pm0.101$\\ \hline
\end{tabular}
\label{t:all_datasets}%
\end{table}%

\section{Conclusion}

We introduced Dose-AIPTB, a nonparametric framework for estimating the
Individual Probability of Treatment Benefit (IPTB) when treatments are
administered at discrete dose levels. By reformulating IPTB estimation as a
binary classification task based on pairwise patient comparisons, our approach
directly addresses the gap between population-level causal inference and
individualized clinical decision-making. The core idea lies in an
attention-based aggregation mechanism that leverages similarity-weighted
comparisons to construct probabilistic benefit estimates while naturally
incorporating dose information. An important property of Dose-AIPTB is that it
can be validated on real data. The corresponding testing set (Validation 1) is
composed of pairs formed from instances in the control and treatment groups.
Since we know the values of $\Delta_{ij}$ for every pair in the testing set,
as well as the probability $p^{+}(\mathbf{z}_{i},\mathbf{x}_{j})$, we can
estimate the model performance.

Comprehensive experiments across six synthetic datasets and the well-known
IHDP benchmark demonstrate that Dose-AIPTB consistently outperforms
established meta-learners (T-learner, S-learner, X-learner, DR-learner) in
ROC-AUC performance. Notably, Dose-AIPTB exhibits remarkable stability under
severe sample imbalance, maintaining consistent performance where baseline
methods degrade substantially. The method's only marginal underperformance
occurred on the Spiral dataset, suggesting that certain complex non-linear
response surfaces remain challenging for pairwise comparison frameworks. Key
limitations include the current focus on discrete rather than continuous
doses, computational complexity scaling with pairwise comparisons, and
reliance on standard causal assumptions. Future work will extend the framework
to continuous dose spaces, integrate uncertainty quantification via conformal
prediction, and develop scalable approximations for large-scale applications.
Applying Dose-AIPTB to high-dimensional covariates and evaluating its clinical
utility through prospective studies represent important translational next
steps. By shifting inference from expected effects to probabilities of
individual benefit while explicitly modeling dose heterogeneity, Dose-AIPTB
offers a principled foundation for dose personalization in precision medicine
with code publicly available at \url{https://github.com/NTAILab/AIPTBDose}.

\begin{credits}
\subsubsection{\ackname} This work is supported by the Russian Science Foundation under grant 25-11-00021.
\end{credits}


\end{document}